\pdfoutput=1

\documentclass[11pt]{article}

\usepackage[final]{acl}

\usepackage{times}
\usepackage{latexsym}
\usepackage{cuted}
\usepackage[T1]{fontenc}

\usepackage[utf8]{inputenc}

\usepackage{microtype}

\usepackage{inconsolata}

\usepackage{graphicx}

\usepackage{amsthm,amsmath,amssymb}
\usepackage{enumitem}
\usepackage{booktabs}       
\usepackage{wrapfig}
\usepackage{subfigure}
\usepackage{multirow}
\usepackage{float}

\usepackage{tcolorbox} 
\tcbuselibrary{theorems} 
\usepackage{xcolor} 

\definecolor{mytheoremfr}{RGB}{165,221,155} 
\definecolor{mytheorembg}{RGB}{226,244,197} 
\newtcbtheorem{Prompt}{Prompt}{colback=mytheorembg!25,colframe=mytheoremfr, boxrule=0.5pt
}{th}

\newcommand{\eg}{\textit{e}.\textit{g}.}

\newcommand{\za}[1]{{\color{black}{#1}}}

\title{Hello Again! LLM-powered Personalized Agent for Long-term Dialogue}

\author{
 \textbf{Hao Li\textsuperscript{1}\thanks{These authors contribute equally to this work.}},
 \textbf{Chenghao Yang\textsuperscript{2}$^*$},
 \textbf{An Zhang\textsuperscript{1}\thanks{An Zhang is the corresponding author.}},
 \textbf{Yang Deng\textsuperscript{3}},
 \textbf{Xiang Wang\textsuperscript{2}},
 \textbf{Tat-Seng Chua\textsuperscript{1}}
\\
 \textsuperscript{1}National University of Singapore
 \\
 \textsuperscript{2}University of Science and Technology of China
 \\
 \textsuperscript{3}Singapore Management University
 \\
 \texttt{18th.leolee@gmail.com}, \texttt{yangchenghao@mail.ustc.edu.cn},
 \\
 \texttt{anzhang@u.nus.edu}, \texttt{ydeng@smu.edu.sg},
 \\
 \texttt{xiangwang123@gmail.com}, \texttt{dcscts@nus.edu.sg}
} 

\begin{document}
\maketitle

\begin{abstract}
Open-domain dialogue systems have seen remarkable advancements with the development of large language models (LLMs). Nonetheless, most existing dialogue systems predominantly focus on brief single-session interactions, neglecting the real-world demands for long-term companionship and personalized interactions with chatbots. Crucial to addressing this real-world need are event summary and persona management, which enable reasoning for appropriate long-term dialogue responses. Recent progress in the human-like cognitive and reasoning capabilities of LLMs suggests that LLM-based agents could significantly enhance automated perception, decision-making, and problem-solving. In response to this potential, we introduce a model-agnostic framework, the \underline{L}ong-term \underline{D}ialogue \underline{Agent} (\textbf{LD-Agent}), which incorporates three independently tunable modules dedicated to event perception, persona extraction, and response generation. For the event memory module, long and short-term memory banks are employed to separately focus on historical and ongoing sessions, while a topic-based retrieval mechanism is introduced to enhance the accuracy of memory retrieval. Furthermore, the persona module conducts dynamic persona modeling for both users and agents. The integration of retrieved memories and extracted personas is subsequently fed into the generator to induce appropriate responses. The effectiveness, generality, and cross-domain capabilities of LD-Agent are empirically demonstrated across various illustrative benchmarks, models, and tasks. The code is released at \url{https://github.com/leolee99/LD-Agent}.

\end{abstract}

\section{Introduction}
\label{sec:introduction}
Open-domain dialogue systems aim to establish long-term, personalized interactions with users via human-like chatbots~\cite{MSC, HAHT, DuLeMon}.
Unlike most existing studies~\cite{DailyDialog, PersonaChat, EmpatheticDialogues} that are limited to brief, single-session interactions spanning 2-15 turns, real-life scenarios often necessitate a chatbot's capability for long-term companionship and familiarity~\cite{MSC, HAHT, CC}.
Achieving this requires the chatbot not only to understand and remember extensive dialogue histories but also to faithfully reflect and consistently update both the user's and its personalized characteristics.

\begin{figure*}[t]
\setlength{\abovecaptionskip}{5pt}   
\setlength{\belowcaptionskip}{0pt}
    \centering    
    \includegraphics[width=0.8\linewidth,trim=50 185 50 185,clip]{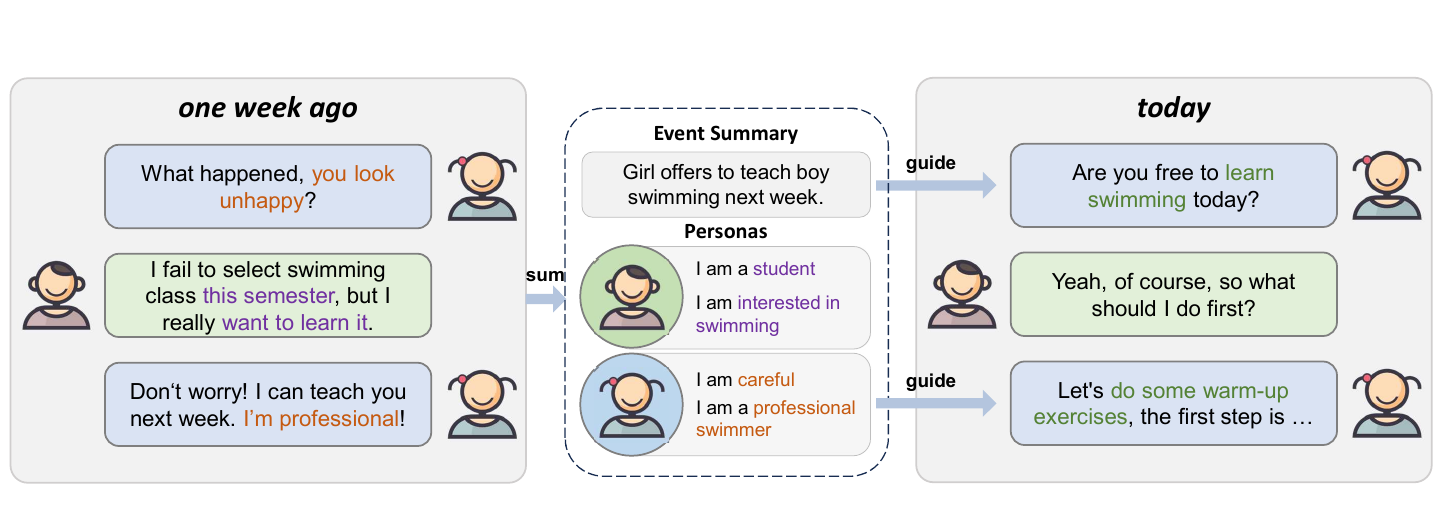}
    \caption{
    The illustration of how event memory and personas guide long-term dialogue. 
    The event summary and personas are extracted from a conversation that occurred one week ago. In today's interaction, the event memory prompts the girl to inquire about the swimming lesson they scheduled last week. The personas, indicating that she is careful and professional in swimming, guide her to offer detailed and professional advice.
    }
    \label{fig:Fig-Intro}
    \vspace{-0.3cm}
\end{figure*}

Motivated by real-life demands, the core challenge of open-domain dialogue systems is to simultaneously maintain long-term event memory and preserve persona consistency~\cite{DIM, D3, UniMC, DuLeMon}.
Existing research often addresses these aspects separately—focusing either on event memory or persona extraction—thereby hindering long-term consistency.
Current strategies for event memory typically involve constructing a memory bank that stores historical event summaries, complemented by retrieval-augmented approaches to access relevant information for response generation~\cite{MemNN, MAD}. 
Studies on persona-based dialogue rang from unidirectional user modeling~\cite{ORIG} to bidirectional agent-user modeling~\cite{PAG, P2, DuLeMon}, enhancing personalized chat abilities by leveraging profile information. 
Worse still, the aforementioned methods are highly dependent on specific model architectures, making them challenging to adapt to other models. Additionally, These dialogue models largely lack zero-shot generalization capabilities, essential for effective deployment across various real-world domains~\cite{HAHT, DuLeMon}.
We conjecture that an optimal long-term dialogue framework should be model-agnostic, deployable in various real-world domains, and capable of autonomously integrating comprehensive data from both event memories and personas, as illustrated in Figure \ref{fig:Fig-Intro}.
However, developing such a model-agnostic, cross-domain, and autonomous framework remains unexplored and challenging.


Benefiting from the excellent human-like cognitive and reasoning abilities of large language models (LLM), there is an increasing trend~\cite{Mind2Web, Voyager, ChatDev, GenerativeAgent, Agent4Rec} to employ LLMs as the cores of agent-based simulation systems to automate the process of perception, decision-making, and problem-solving. While recent studies have developed LLM-powered agents in various fields, such as economics~\cite{SocioDojo}, politics~\cite{WarAgent}, sociology~\cite{Sociology}, and recommendation~\cite{Agent4Rec}, its application in open-domain dialogue remains unexplored. 
To effectively support long-term open-domain dialogue, an LLM-powered dialogue agent framework should exhibit broad generality, cross-domain adaptability, and the ability to dynamically refine information across dimensions like events, user personalities, and agent personalities.


In this paper, we propose \textbf{LD-Agent}—a model-agnostic \textbf{L}ong-term \textbf{D}ialogue \textbf{Agent} framework consisting of three \za{principal} components: an event memory perception module, a persona extraction module, and response generation module \za{(see the framework of LD-Agent in Figure \ref{fig:Framework})}. 
The event memory perception module is designed to enhance coherence across sessions by separately maintaining long-term and short-term memory banks. The long-term memory bank stores vector representations of high-level event summaries from previous sessions, refined through a tunable event summary module. 
The short-term memory bank \za{maintains} contextual information for ongoing conversations.
\za{The persona extraction module, designed to facilitate personalized interactions, incorporates a disentangled, tunable mechanism for accurate user-agent modeling.}
Extracted personas are continuously updated and \za{stored} in a long-term persona bank. 
These personas, along with relevant memories, \za{are then} integrated into the response generation module, guiding the generation of appropriate responses, as depicted in Figure~\ref{fig:Fig-Intro}.

We conduct comprehensive experiments \za{on} two illustrative long-term multi-session daily dialogue datasets, MSC~\cite{MSC} and Conversation Chronicles (CC)~\cite{CC}, to evaluate the effectiveness, generality, and cross-domain capabilities of the proposed framework.
In terms of effectiveness, \za{LD-Agent} achieves state-of-the-art performance on both benchmarks, significantly outperforming existing methods~\cite{HAHT, ChatGLM, Blenderbot}. 
\za{To assess generality, we examine the framework from both model and task perspectives.}
From the model perspective, \za{LD-Agent is evaluated across a range of} both online and offline models, including LLMs~\cite{ChatGLM} and non-LLMs~\cite{Blenderbot}. 
From the task perspective, we extend our evaluation to multiparty dialogue tasks~\cite{GSN}, where LD-Agent also demonstrates substantial improvements, showcasing its adaptability across different models and tasks.
Regarding the method's cross-domain capabilities, we design two cross-domain settings: tuning the model on the MSC dataset and testing it on the CC dataset, and vice versa. 
In both scenarios, \za{LD-Agent} shows competitive performance, \za{nearly matching the results of in-domain training.} 


Our contributions can be summarized as follows:


\begin{itemize}[leftmargin=*,nosep]
    \item We develop LD-Agent, a general long-term dialogue agent framework, considering both historical events, ensuring coherence across sessions and personas, ensuring character consistency.%
    \item We introduce a disentangled, tunable approach for long-term dialogue to ensure the accuracy of each module. The highly modular framework enables it to adapt to various dialogue tasks through module re-training. 
    \item We confirm the superiority of our proposed framework through rigorous experiments across multiple challenging benchmarks, diverse illustrative models, and various tasks. Extensive insightful ablation studies further highlight its effectiveness and generalization.
\end{itemize}

\section{Related Work}
\label{sec:related_work}
\paragraph{Long-term Dialogues.}
Open-domain dialogue aims to develop a human-like chatbot that can emulate human conversation, facilitating free-flowing dialogue on a wide range of topics. However, the dialogue's extent in earlier studies is often limited by conversation length, focusing primarily on brief conversations of about 2-15 turns within a single session~\cite{DailyDialog, PersonaChat, EmpatheticDialogues}. To support more realistic and extended conversations, a series of studies have explored the role of both external~\cite{Prophetic, Commonsense} and internal knowledge~\cite{HAHT, DuLeMon} on maintaining the feasibility of long-term dialogue. Commonly referenced external knowledge, such as commonsense~\cite{Commonsense}, medical~\cite{Medical}, and psychological~\cite{SoulChat} knowledge, serves as supplementary guidance for the reasoning process, ensuring logical coherence in extended contexts. In parallel, internal knowledge captured dynamically during long conversations generally contains historical events~\cite{MSC, GapChat, HAHT, CC} and personas~\cite{DIM, DuLeMon, D3, PAGE}. Historical events are typically summarized and stored into a memory bank to maintain dialogue coherence across sessions, while interlocutors' personas are maintained via a dynamic persona memory bank, ensuring character consistency in long-term conversations. In this study, we focus on the internal knowledge to integrate dynamically updated historical events and personas to conduct long-term personalized conversations.

\paragraph{LLM-based Autonomous Agents.}
AI Agent conception is geared towards autonomous environmental perception, decision-making, and problem-solving capabilities. With the large language models (LLMs) underlining their impressive generalization potential, leading to their widespread adoption as substitutes for human operators in various research fields~\cite{Mind2Web, ChatDev, PsyAgent, Agent4Rec, GenerativeAgent}. Generally, these agents can be categorized into task-oriented agents~\cite{Mind2Web, Voyager, ChatDev, MemoryBank} and simulation-oriented agents~\cite{PsyAgent, Rehearsal, S3, Agent4Rec, RecAgent}. Task-oriented agents are designed to accurately perform predefined tasks, as seen in applications for web assistance~\cite{Mind2Web}, game-playing~\cite{Voyager}, and software development~\cite{ChatDev}. Conversely, simulation-oriented agents are devised to emulate human emotive and cognitive behaviors, having played roles in psychological studies~\cite{PsyAgent}, social networking platforms~\cite{S3}, conflict resolution scenarios~\cite{Rehearsal}, and recommendation systems~\cite{Agent4Rec, RecAgent}. In addition, recent progress has seen the advent of individual-level agents that are utilized to simulate specific character behaviors, enhancing the realism and personalization of user-agent interactions~\cite{CharacterLLM, CharacterGLM, RoleLLM}. This paper falls into simulation-oriented agents to build a human-like open-domain dialogue agent with memory retrieval and character analysis modules.
\section{Method}
\label{sec:method}
In this section, we introduce the LD-Agent in detail with the framework shown in Figure~\ref{fig:Framework}. We first introduce the task definition of long-term dialogue in Section.~\ref{sec:task}. Consequently, we separately introduce the mechanism of event perception (Section.~\ref{sec:event}), dynamic personas extraction (Section.~\ref{sec:persona}), and response generation (Section.~\ref{sec:response}).

\subsection{Task Definition}
\label{sec:task}
\za{
The goal of the long-term multi-session dialogue task is to generate an appropriate response $r$, by utilizing the context of the current session $C$, along with selected information extracted from historical session $H$.
In this task, the current conversation session $C$ is defined as $\{u_1,u_2,\dots,u_{d_c-1},u_{d_c}\}$, where each $u_i$ represents $i$-th utterance, and $d_c$ represents $d_c$ turns of the current session.
Each historical session within $H$ in $N$ historical sessions is denoted as $H^i$, containing $\{h^i_1, h^i_2, \dots, h^i_{d_i}\}$, where $d_i$ is the number of utterances of the $i$-th conversational session.
Distinct from single-session dialogue models, a long-term multi-session dialogue system integrates both current and long-term historical conversational cues to generate contextually appropriate responses. 
}


\subsection{Event Perception}
\label{sec:event}

\begin{figure*}[t]
\setlength{\abovecaptionskip}{5pt}   
\setlength{\belowcaptionskip}{0pt}
    \centering    \includegraphics[width=0.75\textwidth,trim=160 140 160 140,clip]{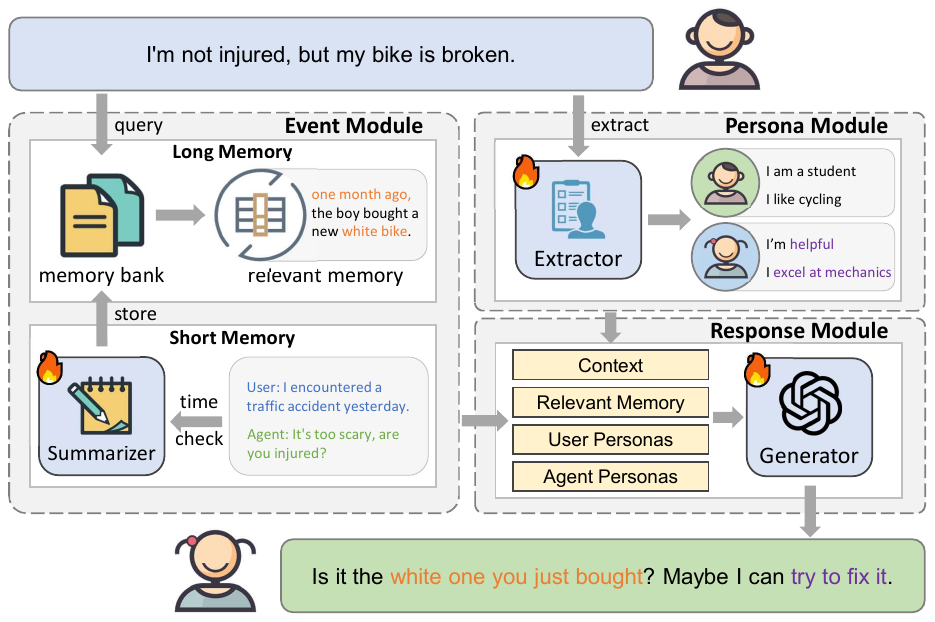}
    \caption{The Framework of LD-Agent. The event module stores historical memories from past sessions in long-term memory and current context in short-term memory. The persona module dynamically extracts and updates personas for both users and agents from ongoing utterances, storing them in a persona bank for each character. The response module then synthesizes this data to generate informed and appropriate responses.}
    \label{fig:Framework}
\vspace{-0.3cm}
\end{figure*}

The event memory module is designed to perceive historical events to generate coherent responses across intervals. 
In Figure~\ref{fig:Framework}, 
\za{this event memory module is divided into two sub-modules that focus separately on long-term and short-term memory.}

\subsubsection{Long-term Memory} 

\paragraph{Memory Storage.} 
The long-term memory module aims to \za{extract and encode} events from past sessions.
\za{Specifically, this involves recording the occurrence times $t$ and brief summaries $o$ into representations that are stored in a low-cost memory bank $M_L=\{\phi(t_j, o_j) \mid j \in \{1,2, \dots, l\}\}$.
Here, $\phi(\cdot)$ indicates the text encoder (\eg, MiniLM~\cite{MiniLM}), and $l$ specifies the length of the memory bank.
The encoded representations are then efficiently retrieved through an embedding-based mechanism, which enhances the accessibility of the stored memory. 
}


\paragraph{Event Summary.}
\label{sec:summary_details}
Different from previous agent approaches~\cite{GenerativeAgent, Agent4Rec, MemoryBank} that entirely rely on LLM's zero-shot ability to excavate and summarize events, we apply instruction tuning~\cite{IT} to the event summary module, which can directly improve the event summary quality. 
Specifically, we rebuild the DialogSum dataset~\cite{DialogSum}, a large-scale dialogue summarization dataset, into the following format: (1) an introduction to the task background, (2) the related conversations that need to be understood, and (3) detailed summarization requests. 
These three parts serve as input prompts (see Appendix.~\ref{event_summary_prompt} for more details), combined with the original summaries from DialogSum as answers, and are jointly used to fine-tune the event summary module, \za{thereby directly improving the quality of event summarization.}

\paragraph{Memory Retrieval.} 
\label{section:memory_retrieval}
To improve retrieval accuracy, we employ a retrieval mechanism that comprehensively considers semantic relevance, topic overlap, and time decay. 
Optimizing the retrieval accuracy of agent memory is challenging due to the difficulty in obtaining accurate memory retrieval data. 
Most existing methods~\cite{GenerativeAgent, Agent4Rec} use event summaries as keys and context as queries, calculating the query-key semantic relevance score $s_{\text{sem}}$ to find relevant memories, which inevitably results in significant errors. 
To enhance retrieval reliability, we extract nouns from corresponding conversations with the summaries to construct a topic library $V$ and calculate topic overlap score $s_{\text{top}}$ by:
\begin{equation}
    s_{\text{top}}=\frac{1}{2}(\frac{\left | V_q\cap V_k \right |}{\left | V_{q} \right |}+\frac{\left | V_q\cap V_k \right |}{\left | V_{k} \right |}),
\label{eq:overlap_compute}
\end{equation}
where $V_q$, $V_k$ denote the topic noun set of query and key. Additionally, we refer to the recency coefficient used by~\citet{GenerativeAgent} and apply an exponential decay function as time decay coefficient $\lambda_t=e^{{-t}/{\tau}}$ to reweight the overall retrieval score $s_r$, signified as Eq~\ref{eq:retrieval_score}. $\tau$ is a temperature coefficient set to 1e+7 in our setting.
\begin{equation}
    s_{\text{overall}} = \lambda_t (s_{\text{sem}} + s_{\text{top}}).
\label{eq:retrieval_score}
\end{equation}
To avoid retrieving inappropriate memory due to no suitable memories existing, we implement a semantic threshold $\gamma$ set to 0.5 in our setting. Only memories with semantic score $s_\text{sem}$ greater than $\gamma$ could be retrieved. If no appropriate memories are retrieved, ``No relevant memory'' will be returned. 
Eventually, the process of retrieving relevant memory can be denoted as $m = \psi(M_L, \gamma)$.

\subsubsection{Short-term Memory} 
The short-term memory module actively manages a dynamic dialogue cache $M_S=\{(t_i,u_i)|i=\{1,2,3,\dots,r_c\}\}$ with timestamps to preserve the detailed context of the current session. 
Upon receiving a new utterance $u'$, the module first evaluates the time interval between the current time $t'$ and the last recorded time $t_{r_c}$ in the cache. 
If this interval exceeds a threshold $\beta$ (600 seconds in our setting), the module triggers the long-term memory module to summarize the cached dialogue entries, creating new event records for storage in the long-term memory bank. 
Simultaneously, the short-term memory cache is cleared, and the new dialogue record $(t',u')$ is added to the cache. 
\za{The mathematical expression of this process is given by:}
\begin{equation}
    \begin{aligned}
    M'_L &= M_L \cup \{(\phi(t_{r_c}, A(M_S))\},\\
    M_S &= \{(t',u')\}.
    \end{aligned}
\label{eq:memory_update}
\end{equation}
where $M'_L$ denotes the updated long-term memory bank, $o=A(\cdot)$ is the event summary function, \za{which process the accumulated dialogue in $M_S$.}

\subsection{Dynamic Personas Extraction}
\label{sec:persona}
The persona module is pivotal in maintaining long-term persona consistency for both participants in a dialogue system. 
Drawing inspiration from prior work~\cite{DuLeMon}, we adopt a bidirectional user-agent modeling approach, utilizing a tunable persona extractor to manage long-term persona bank $P_u$ and $P_a$ for the user and agent, respectively. 
Specifically, we develop an open-domain, utterance-based persona extraction dataset derived from MSC~\cite{MSC}. 
We enhance the persona extractor with LoRA-based instruction tuning, which allows for the dynamic extraction of personality traits during conversations. 
These traits are subsequently stored in the corresponding character's persona bank. 
For utterances devoid of personality traits, the module outputs ``No Trait''. 
Additionally, we employ a tuning-free strategy that harnesses the zero-shot capabilities of LLM models to directly extract personas based on prompts (see Appendix.~\ref{persona_extraction_prompt}).  
To further improve the ability to excavate user personas without training, we adjust our reasoning strategy from direct reasoning to a Chain-of-Thought reasoning~\cite{CoT}.

\subsection{Response Generation}
\label{sec:response}
Upon receiving a new user utterance $u'$, the agent \za{integrates various inputs:} retrieved relevant memories $m$, short-term context $M_S$, and the personas $P_u$ and $P_a$ for the user and agent, respectively. 
\za{These combined inputs} are fed into a response generator to deduce an appropriate response \za{$r$}, formulated as
\begin{equation}
r=G(u',m,M_S,P_u,P_a).
\label{eq:response_generation}
\end{equation}
To enhance the agent's ability for coherent and contextually appropriate responses,
we develop a long-term, multi-session dialogue dataset, featuring dynamic retrieval memories, context, and personas sourced from the MSC and CC datasets for generator tuning. 
Specifically, for each sample, covering five sessions, we dynamically simulate the entire progression of the conversation. 
As each new utterance is introduced, the previously tuned modules for event summarization, persona extraction, and memory retrieval are utilized to collect the necessary context, retrieved memories, and both user and agent personas related to the utterance.
This comprehensive data is then integrated into a response generation prompt (see Appendix.~\ref{response_generation_prompt}). 
The original responses from the MSC and CC datasets are used as ground truth sentences.
\section{Experiments}
\label{sec:experimetns}
We aim to answer the following research questions:
\begin{itemize}[leftmargin=*,nosep]
   \item \textbf{RQ1:} How does LD-Agent perform in long-term dialogue tasks?
   \item \textbf{RQ2:} How is the generality and practicality of LD-Agent?

\end{itemize}

\subsection{Evaluation Settings}
In this subsection, we briefly introduce the experimental dataset, evaluation metrics, and baseline models in our study. Detailed evaluation settings are elaborated in Appendix.~\ref{sec:detailed_eval_settings}.

\begin{table*}[h]
\centering
\captionsetup{skip=5pt}
  \renewcommand{\arraystretch}{0.8}
\resizebox{0.8\textwidth}{!}{%
\begin{tabular}{@{}clccc|ccc|ccc|ccc@{}}
\toprule
 & & \multicolumn{3}{c}{\textbf{Session 2}} & \multicolumn{3}{c}{\textbf{Session 3}} & \multicolumn{3}{c}{\textbf{Session 4}} & \multicolumn{3}{c}{\textbf{Session 5}} \\
\cmidrule(lr){3-5} \cmidrule(lr){6-8} \cmidrule(lr){9-11} \cmidrule(lr){12-14}
\multicolumn{1}{c}{\multirow{-2}{*}{\textbf{}}} & \multicolumn{1}{c}{\multirow{-2}{*}{\textbf{Model}}} & \textbf{BL-2} & \textbf{BL-3} & \textbf{R-L} & \textbf{BL-2} & \textbf{BL-3} & \textbf{R-L} & \textbf{BL-2} & \textbf{BL-3} & \textbf{R-L} & \textbf{BL-2} & \textbf{BL-3} & \textbf{R-L} \\
\midrule
\multicolumn{14}{c}{\textbf{MSC}}\\
\midrule
& ChatGLM & 5.44 & 1.49 & 16.76 & 5.18 & 1.55 & 15.51 & 5.63 & 1.33 & 16.35 & 5.92 & 1.45 & 16.63 \\
& ChatGLM\textsubscript{LDA} & 5.74 & 1.73 & 17.21 & 6.05 & 1.73 & 16.97 & 6.09 & 1.59 & 16.76 & 6.60 & 1.94 & 17.18 \\
& ChatGPT & 5.22 & 1.45 & 16.04 & 5.18 & 1.55 & 15.51 & 4.64 & 1.32 & 15.19 & 5.38 & 1.58 & 15.48 \\
\multirow{-4}{*}{\textbf{Zero-shot}} & ChatGPT\textsubscript{LDA} & 8.67 & 4.63 & 19.86 & 7.92 & 3.55 & 18.54 & 7.08 & 2.97 & 17.90 & 7.37 & 3.03 & 17.86 \\
\midrule
 & HAHT & 5.06 & 1.68 & 16.82 & 4.96 & 1.50 & 16.48 & 4.75 & 1.45 & 15.82 & 4.99 & 1.51 & 16.24 \\
 & BlenderBot & 5.71 & 1.62 & 16.15 & 8.10 & 2.50 & 18.23 & 7.55 & 1.96 & 17.45 & 8.02 & 2.36 & 17.65 \\
 & BlenderBot\textsubscript{LDA} & 8.45 & 3.27 & 19.07 & 8.68 & 3.06 & 18.87 & 8.16 & 2.77 & 18.06 & 8.31 & 2.69 & 18.19 \\
 & ChatGLM & 5.48 & 1.59 & 17.65 & 6.12 & 1.78 & 17.91 & 6.14 & 1.63 & 17.78 & 6.16 & 1.69 & 17.65 \\
  & ChatGLM\textsubscript{LDA} & 7.42 & 2.46 & 20.04 & 7.47 & 2.40 & 19.50 & 7.52 & 2.32 & 19.55 & 7.36 & 2.37 & 19.16 \\
\multirow{-6}{*}{\textbf{Tuning}}& $\text{ChatGLM}_\text{LDA}^*$ & \textbf{10.70} & \textbf{5.63} & \textbf{23.31} & \textbf{10.03} & \textbf{5.12} & \textbf{21.55} & \textbf{9.07} & \textbf{4.06} & \textbf{20.19} & \textbf{8.96} & \textbf{4.01} & \textbf{19.94} \\
\midrule
\multicolumn{14}{c}{\textbf{CC}}\\
\midrule
\multirow{4}{*}{\textbf{Zero-shot}} & ChatGLM & 8.94 & 4.44 & 21.54 & 8.34 & 4.03 & 21.00 & 8.28 & 3.82 & 20.67 & 8.12 & 3.81 & 20.54 \\
 & ChatGLM\textsubscript{LDA} & 9.53 & 4.82 & 22.76 & 9.22 & 4.43 & 22.18 & 9.15 & 4.48 & 22.18 & 8.99 & 4.43 & 22.10 \\
 & ChatGPT & 10.57 & 5.50 & 22.10 & 10.58 & 5.59 & 22.04 & 10.61 & 5.58 & 21.92 & 10.17 & 5.22 & 21.45 \\
 & ChatGPT\textsubscript{LDA} & 15.89 & 11.01 & 26.96 & 12.92 & 8.27 & 24.31 & 12.20 & 7.35 & 23.69 & 11.54 & 6.74 & 22.87 \\
\midrule
\multirow{4}{*}{\textbf{Tuning}}  & HAHT & 11.59 & 6.20 & 24.09 & 11.52 & 6.14 & 23.94 & 11.27 & 5.99 & 23.77 & 10.69 & 5.51 & 23.04 \\
& BlenderBot & 8.99 & 4.86 & 21.58 & 9.44 & 5.19 & 22.13 & 9.46 & 5.21 & 22.08 & 8.99 & 4.75 & 21.73 \\
 & BlenderBot\textsubscript{LDA} & 14.47 & 10.16 & 27.91 & 15.66 & 11.33 & 29.10 & 15.13 & 10.80 & 28.38 & 14.08 & 9.72 & 27.37 \\
 & ChatGLM & 15.89 & 9.90 & 30.59 & 15.97 & 10.06 & 30.27 & 16.10 & 10.31 & 30.54 & 15.10 & 9.34 & 29.43 \\
 & ChatGLM\textsubscript{LDA} & \textbf{25.69} & \textbf{19.53 }& \textbf{39.67} & \textbf{25.93} & \textbf{19.72} & \textbf{39.15} & \textbf{25.82} & \textbf{19.40} & \textbf{39.05} & \textbf{24.26} & \textbf{18.16} & \textbf{37.61} \\
\bottomrule
\end{tabular}
}
\caption{Experimental results of the automatic evaluation for response generation on MSC and CC. * denotes using annotations as personas.}
\label{table:msc_cc}
\vspace{-2mm}
\end{table*}

\paragraph{Datasets.} To investigate the effectiveness of LD-Agent on long-term dialogue scenarios, extensive experiments are conducted on two illustrative multi-session datasets, \textbf{MSC}~\cite{MSC} and \textbf{CC}~\cite{CC}, each comprising 5 sessions with approximately 50 conversational turns per sample. The experiments cover model independence assessment, module ablation, persona extractor analysis, and cross-domain evaluation. Additionally, to evaluate the transferability of the LD-Agent, we apply our method to the \textbf{Ubuntu IRC benchmark}~\cite{GSN}, a dataset known for its multiparty interaction tasks.


\paragraph{Metrics.} 
Our evaluation combines both automatic and human assessments to thoroughly investigate the effectiveness of LD-Agent. For automatic evaluation, we use three widely used standard metrics: BLEU-N (BL-N)~\cite{BLEU}, ROUGE-L (R-L)~\cite{ROUGE}, and METEOR (MET)~\cite{METERO} to measure the quality of response generation. Additionally, accuracy (ACC) is employed to evaluate the classification performance of the persona extractor. In human evaluation, we measure topic coherence across sessions, interaction fluency, and user engagement using the metrics of coherence, fluency, and engagingness, respectively.



\paragraph{Baselines.}
To demonstrate the effectiveness and model independence of LD-Agent, we deploy it on multiple platforms and models. Specifically, the LLM-based models (online model: ChatGPT; offline model: ChatGLM~\cite{ChatGLM}) and traditional language models (BlenderBot~\cite{Blenderbot}, and BART~\cite{BART}) are employed as our baselines. In our experiments, The notation ``\textbf{Model}\textsubscript{LDA}'' denotes models that incorporate the LD-Agent framework, while ``\textbf{Model}'' refers to the original baseline models without LD-Agent. Additionally, we also utilize HAHT~\cite{HAHT}, the previous state-of-the-art model in long-term dialogue task, as a contrast.

\subsection{Evaluation Pipeline}
\paragraph{Automatic Evaluation Pipeline.}
For the automatic evaluation, We first utilize the first session in MSC or CC to initialize the conversation. Afterward, we calculate generation accuracy for each utterance. For instance, in a 30-turn dialogue where 15 utterances come from Speaker B, who will be role-played by the Agent. Accuracy is calculated based on all of these 15 utterances. To simulate a realistic dialogue scenario, before generating each utterance, we first input all the preceding ground-truth conversations into the LD-Agent framework to simulate the real historical interaction process. During the simulation, personas and memories are automatically extracted to assist in generating the current utterance. Additionally, since the MSC dataset has annotated personas, we also evaluated using these annotations as personas instead of extracting them, marked with *.

\paragraph{Human Evaluation Pipeline.}
\label{add: human_details}
In Section~\ref{sec:human_eval}, we conduct human evaluation on memory retrieval and response generation. Specifically, we employ 8 students to assist with the assessment and evaluate LD-Agent on two tasks: memory retrieval evaluation and  response generation evaluation. The detailed guidelines of human evaluation are shown in Figure~\ref{fig:human_guidelines} of Appendix.

\subsection{Results of Multi-Session Dialogue}

We adopt two multi-session dialogue dataset to evaluate our method in long-term dialogue scenarios. The first session is used to initialize conversation and the subsequent four sessions are used to evaluate the performance of long-term dialogue. To ensure consistency with real-world scenarios, we simulate all previous dialogues from the start before calculating each utterance's generation accuracy.
In these experiments, LD-Agent is applied to both zero-shot models, including ChatGLM and ChatGPT, and to tuned models like BlenderBot and ChatGLM with the results reported in Table~\ref{table:msc_cc}.
\paragraph{Impressive performance on long-term dialogue tasks.} On both datasets, all models using LD-Agent consistently achieve significant improvements across all sessions and metrics, showcasing the powerful ability of LD-Agent on supporting long-term dialogue. Most notably, compared to previous state-of-the-art model HAHT, BlenderBot employing LD-Agent with similar parameter scale to HAHT, outperforms it with a large performance gap on BLEU-2 ranging from session 2 to 5 on both datasets. This further highlights the effectiveness of LD-Agent.

\paragraph{Remarkable generality of LD-Agent.} The generality of LD-Agent are proved from two aspects: data transferability and model transferability. The consistently improvements brought by LD-Agent on both benchmarks demonstrate the generality of our framework on various long-term dialogue scenarios. In parallel, we observe that LD-Agent also plays positive roles in the zero-shot setting, employing to the online model of ChatGPT and the offline model of ChatGLM. In the tuning setting, LD-Agent achieves significant enhancements on both LLM of ChatGLM and traditional model of BlenderBot, fully proving the remarkable model transferability of LD-Agent. These results comprehensive demonstrate the generality of LD-Agent. 

\begin{table*}[h]
\centering
\captionsetup{skip=5pt}
  \renewcommand{\arraystretch}{0.8}
\resizebox{0.8\textwidth}{!}{%
\begin{tabular}{@{}clccc|ccc|ccc|ccc@{}}
\toprule
 & & \multicolumn{3}{c}{\textbf{Session 2}} & \multicolumn{3}{c}{\textbf{Session 3}} & \multicolumn{3}{c}{\textbf{Session 4}} & \multicolumn{3}{c}{\textbf{Session 5}} \\
\cmidrule(lr){3-5} \cmidrule(lr){6-8} \cmidrule(lr){9-11} \cmidrule(lr){12-14}
\multicolumn{1}{c}{\multirow{-2}{*}{\textbf{}}} & \multicolumn{1}{c}{\multirow{-2}{*}{\textbf{Model}}} & \textbf{BL-2} & \textbf{BL-3} & \textbf{R-L} & \textbf{BL-2} & \textbf{BL-3} & \textbf{R-L} & \textbf{BL-2} & \textbf{BL-3} & \textbf{R-L} & \textbf{BL-2} & \textbf{BL-3} & \textbf{R-L} \\
\midrule
 & Baseline & 5.48 & 1.59 & 17.65 & 6.12 & 1.78 & 17.91 & 6.14 & 1.63 & 17.78 & 6.16 & 1.69 & 17.65 \\
 & + Mem & 7.57 & 2.49 & 19.50 & 7.70 & 2.48 & 19.46 & 7.53 & 2.31 & 19.26 & 7.56 & 2.33 & 19.03 \\
 & + Persona\textsubscript{ user} & 7.54 & 2.57 & 19.68 & 7.51 & 2.38 & 19.39 & 7.30 & 2.09 & 18.80 & 7.08 & 2.27 & 18.79 \\
 & + Persona\textsubscript{ agent} & 7.00 & 2.27 & 18.70 & 7.23 & 2.33 & 18.75 & 7.32 & 2.18 & 18.47 & 7.13 & 2.36 & 18.48 \\
 & Full & \textbf{10.70} & \textbf{5.63} & \textbf{23.31} & \textbf{10.03} & \textbf{5.12} & \textbf{21.55} & \textbf{8.96} & \textbf{4.01} & \textbf{19.94} & \textbf{9.07} & \textbf{4.06} & \textbf{20.19} \\
\bottomrule
\end{tabular}
}
\caption{Ablation study results of LD-Agent on MSC. The experiments are conducted on tuned ChatGLM. 
Baseline denotes the model tuned with context of current session. ``+ module name'' indicates the model tuned solely with context and corresponding module. ``Full'' indicates the model tuned with all modules.} 
\label{table:msc_ablation}
\vspace{-0.3cm}
\end{table*}

\subsection{Ablation Studies}
\label{sec:ablation}
To further analyze the effectiveness of each components, we conduct ablation studies for memory module and personas module. We adopt ChatGLM as our backbone, which is tuned solely using the context of the current session, referred to here as ``Baseline''. Afterward, we separately add ``Event Memory'', ``Agent personas'', and ``User personas'' modules for additional tuning on top of the baseline. The results are presented in Table~\ref{table:msc_ablation}.

The results clearly prove that all modules positively influence long-term dialogue capabilities, with the event memory module contributing the most. It is worth noting that although all modules experience a performance decline as sessions increase, the addition of the event memory module results in more stable performance compared to the use of user or agent personas. This highlights the critical role of event memory in maintaining coherence across multiple sessions.


\begin{table}[h]
\centering
\captionsetup{skip=5pt}
  \renewcommand{\arraystretch}{0.9}
\resizebox{0.48\textwidth}{!}{%
\begin{tabular}{@{}clcccc|ccc@{}}
\toprule
 & & \multicolumn{4}{c|}{\textbf{Extraction}} & \multicolumn{3}{c}{\textbf{Generation}} \\
\cmidrule(lr){3-6} \cmidrule(lr){7-9}
\multicolumn{1}{c}{\multirow{-2}{*}{\textbf{}}} & \multicolumn{1}{c}{\multirow{-2}{*}{\textbf{Extractor}}} & \textbf{BL-2} & \textbf{BL-3} & \textbf{R-L} & \textbf{ACC} & \textbf{BL-2} & \textbf{BL-3} & \textbf{R-L}  \\
\midrule
& CoT & 5.05 & 2.69 & 25.54 & 61.6 & 5.82 & 1.69 & 16.95 \\
& Tuning & \textbf{8.31} & \textbf{5.65} & \textbf{43.70} & \textbf{77.8} & \textbf{6.12} & \textbf{1.75} & \textbf{17.03} \\

\bottomrule
\end{tabular}
}
\caption{The effect of different extractors on persona extraction and response generation on MSC.}
\label{table:persona}
\vspace{-0.5cm}
\end{table}

\subsection{Persona Extraction Analysis}
To explore the effect of different persona extractor, including zero-shot ChatGLM with Chain-of-Thought~\cite{CoT} and ChatGLM tuned on the persona extraction dataset collected from MSC training set, we carry out comparison experiments on two perspectives: Persona Extraction Accuracy and Impact to Response Generation. The results are shown in Table~\ref{table:persona}.

\paragraph{Extraction Accuracy.}

We evaluate the extraction accuracy on the persona extraction dataset collected from MSC testing set, through BLEU-2/3, R-L, and ACC. ACC is to assess the classification accuracy of dividing utterance into ``with personas'' or ``without personas''. The results of extraction in Table~\ref{table:persona} show that the extractor after tuning performs better than CoT on all metrics. The higher BLEU and R-L indicates the tuned extractor performs better capability to extract personas, while higher ACC indicates a stronger capability to distinguish whether personas are contained in an utterance.

\paragraph{Impact to Response Generation.}
In addition, to explore the effect of different persona extractor to the final response generation, we conduct experiments on MSC by comparing the results of zero-shot ChatGLM\textsubscript{LDA} with personas extracted by CoT and tuned extractor, respectively. The Generation results in Table~\ref{table:persona} indicate the tuned extractor performs better in most sessions. As the number of sessions increases, the gap is also constantly expanding, demonstrating tuned extractor is more suitable for long-term dialogue.



\begin{figure}[h]
\setlength{\abovecaptionskip}{5pt}   
\setlength{\belowcaptionskip}{0pt}
    \centering
    \subfigure[Retrieval Mechanism]{
        \includegraphics[width=0.45\linewidth,trim=0 10 0 10,clip]{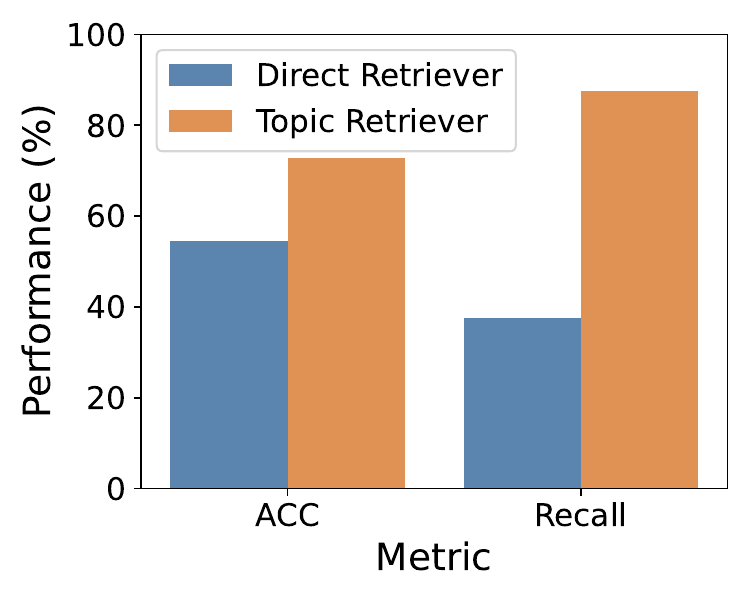}
        \label{fig:retrieval}
    }
    \subfigure[Response Generation]{
        \includegraphics[width=0.45\linewidth,trim=0 10 0 10,clip]{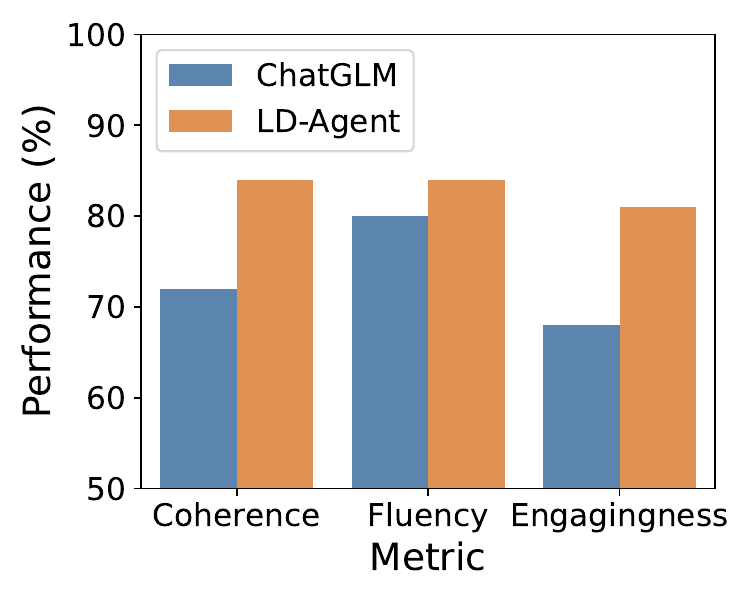}
        \label{fig:generation}
    }
    \caption{The results of human evaluation on retrieval mechanism and response generation.} 
    \label{fig:human}
    \vspace{-0.5cm}
\end{figure}

\subsection{Human Evaluation}
\label{sec:human_eval}
To further explore the performance of LD-Agent in real-life conversation, we employ 8 students to assist with the assessment on memory recall and response generation according to the guidelines in Figure~\ref{fig:human_guidelines}. More human evaluation details can be found in Appendix.~\ref{add: human_details}.

\paragraph{Retrieval Mechanism Analysis.} Retrieval mechanism plays a crucial role for event memory accurately utilized in long-term dialogue. To evaluate the superiority of topic-based retrieval than direct semantic retrieval commonly used in previous methods, we conduct an event memory human evaluation. We first initialize a conversation using first four sessions and store event memories for each session into long-term memory bank. In the last session, we let evaluators select relevant memories from long-term memory bank for each utterance as the ground truths. Consequently, we separately utilize direct semantic retrieval and topic-based retrieval to search relevant memories for each utterance, and calculate the accuracy and recall based on human annotations. The results are shown in Figure~\ref{fig:retrieval}. The topic-based retrieval outperforms direct semantic retrieval with significant gap on both ACC and Recall, proving that our retrieval method accurately retrieves relevant memories.

\paragraph{Response Generation Analysis.}
To further validate the superiority of LD-Agent in long-term open-domain dialogue tasks, we organize multiple multi-session human-bot conversations on ChatGLM with LD-Agent and w/o LD-Agent. We first initialize a predefined dialogue as the first session for all chatbots. Subsequently, we employ some human evaluators to chat with each chatbot with a time interval from first session. We ask human evaluators engage in 2-3 chat sessions with both the original ChatGLM and LD-Agent according to their own thoughts. Afterward, the interactions will be evaluated on three aspects: coherence, fluency and engagingness. The results in Figure~\ref{fig:generation} demonstrate the advantages of LD-Agent in long-term real-life dialogue scenarios.

\begin{table}[h]
    \centering
    \captionsetup{skip=5pt}
      \renewcommand{\arraystretch}{0.8}
    \resizebox{0.48\textwidth}{!}{%
    \begin{tabular}{@{}lccc|ccc@{}}
    \toprule
    & \multicolumn{3}{c}{\textbf{Session 2}} & \multicolumn{3}{c}{\textbf{Session 3}} \\
    \cmidrule(lr){2-4} \cmidrule(lr){5-7}
    \multicolumn{1}{c}{\multirow{-2}{*}{\textbf{Model}}} & \textbf{BL-2} & \textbf{BL-3} & \textbf{R-L} & \textbf{BL-2} & \textbf{BL-3} & \textbf{R-L} \\
    \midrule
    Zero-shot & 9.53 & 4.82 & 22.76 & 9.22 & 4.43 & 22.18 \\
    Zero-shot\textsubscript{LDA} & 8.94 & 4.44 & 21.54 & 8.34 & 4.03 & 21.00 \\
    MSC-tuning & 8.37 & 3.88 & 22.93 & 8.49 & 3.99 & 22.96 \\
    MSC-tuning\textsubscript{LDA} & 21.71 & 15.42 & 34.97 & 20.87 & 14.74 & 34.01 \\
    CC-tuning & 15.89 & 9.90 & 30.59 & 15.97 & 10.06 & 30.27 \\
    CC-tuning\textsubscript{LDA} & \textbf{25.69} & \textbf{19.53} & \textbf{39.67} & \textbf{25.93} & \textbf{19.72} & \textbf{39.15} \\
    \midrule
    & \multicolumn{3}{c}{\textbf{Session 4}} & \multicolumn{3}{c}{\textbf{Session 5}} \\
    \cmidrule(lr){2-4} \cmidrule(lr){5-7}
    Zero-shot & 9.15 & 4.48 & 22.18 & 8.99 & 4.43 & 22.10 \\
    Zero-shot\textsubscript{LDA} & 8.28 & 3.82 & 20.67 & 8.12 & 3.81 & 20.54 \\
    MSC-tuning & 7.97 & 3.75 & 22.15 & 7.60 & 3.70 & 21.87 \\
    MSC-tuning\textsubscript{LDA} & 19.57 & 13.51 & 32.72 & 18.59 & 12.80 & 31.68 \\
    CC-tuning & 16.10 & 10.31 & 30.54 & 15.10 & 9.34 & 29.43 \\
    CC-tuning\textsubscript{LDA} & \textbf{25.82} & \textbf{19.40} & \textbf{39.05} & \textbf{24.26} & \textbf{18.16} & \textbf{37.61} \\
    \bottomrule
    \end{tabular}
    }
    \caption{The results of cross-domain evaluation on CC. ``Zero-shot'' indicates the ChatGLM without tuning. ``CC-tuning'' indicates the ChatGLM tuned on CC. ``MSC-tuning'' indicates the ChatGLM tuned on MSC.}
    \label{table:cross_domain}
\end{table}
\vspace{-5mm}
\subsection{Generality Analysis}
\label{sec: cross}
We further explore its generality from two perspectives: cross-domain and cross-task capability.

\paragraph{Cross-domain Results.}
The cross-domain capability is crucial for open-domain dialogue task. Poor cross-domain performance, common in models tuned with specific datasets, limits their real-world practicality. To assess our tuned model's real-world potential, we conduct cross-evaluation on MSC and CC, two datasets with significant domain gaps due to different collection methods, including manual annotation and LLM generation. We first tune ChatGLM on MSC and test it on CC, then reverse the process. We show the results on CC in Table~\ref{table:cross_domain}, and the full results on MSC and CC in Table~\ref{table:cross_domain_both} of Appendix. It shows that models tuned on one dataset still performs well on the other dataset, only with a slight performance decrease than the models tuned on the same dataset. Besides, cross-domain tuned models consistently outperform zero-shot models by a significant margin. These experiments highlight strong cross-domain and practical potential of LD-Agent.
\begin{table}[h]
\setlength{\tabcolsep}{3mm}
\centering
\captionsetup{skip=5pt}
  \renewcommand{\arraystretch}{0.8}
\scalebox{0.65}{
\begin{tabular}{@{}lccccccc@{}}
\toprule
Model                          & \textbf{BL-1} & \textbf{BL-2} & \textbf{BL-3} & \textbf{BL-4} & \textbf{MET} & \textbf{R-L} \\ 
\midrule
GPT-2 & 10.37  & 3.60   & 1.66   & 0.93   & 4.01   & 9.53    \\
GSN & 10.23  & 3.57   & 1.70   & 0.97   & 4.10   & 9.91    \\
HeterMPC\textsubscript{BART} & 12.26  & 4.80   & \textbf{2.42}   & \textbf{1.49}   & 4.94   & 11.20   \\
BART & 11.25   & 4.02   & 1.78   & 0.95   & 4.46   & 9.90  \\
\midrule
BART\textsubscript{LDA} & \textbf{14.40}   & \textbf{4.92}   & 2.07   & 1.00   & \textbf{5.30}   & \textbf{12.28}  \\
\bottomrule
\end{tabular}}
\caption{Multi-party performance on the Ubuntu IRC.}
\label{table:ubuntu}
\end{table}
\vspace{-5mm}
\paragraph{Cross-task Results.}
The other capability worth exploring is the transferability of LD-Agent to different dialogue tasks. We explore the effectiveness of our method on multiparty dialogue, a task requires playing multiple roles simultaneously. We conduct experiments on Ubuntu IRC dataset~\cite{GSN}, a commonly used multiparty dialogue dataset. where our backbone adopts BART~\cite{BART}. We compare our method with some previous multiparty dialogue methods, including GPT-2~\cite{GPT-2}, GSN~\cite{GSN}, HeterMPC\textsubscript{BART}~\cite{HeterMPC}, and BART tuned without prompt. The results are reported at Table~\ref{table:ubuntu}. It can be seen that BART tuned with LD-Agent obtained the state-of-the-art performance in most metrics, outperforming previous multiparty dialogue approach HeterMPC\textsubscript{BART}, which also employs BART as backbone. This well proves the powerful task transferability of LD-Agent.


\section{Conclusion}
\label{sec:conclusion}
In this work, we delved into the long-term open-domain dialogue agent to meet the real-life chatbot demands for long-term companionship and personalized interactions. Unlike most prior solutions, which primarily focus on brief conversations spanning 2-15 turns within a single session, long-term dialogue agents could support consistent interactions over extended periods, even with significant time gaps between sessions. These agents can also perceive and adapt to user personalities, enabling them to deliver more accurate and personalized services. To achieve this, we introduced a general long-term dialogue agent framework, LD-Agent, which comprehensively considers both historical events and user-agent personas to support coherent and consistent conversation. Its decomposition into three learnable modules significantly enhances its adaptability and transferability. Extensive experiments demonstrated LD-Agent's strong capability in long-term dialogue tasks, showing its practicality across multiple benchmarks, models, and tasks.

\section*{Limitations}
Though LD-Agent exhibits impressive effectiveness and generality on long-term dialogue, we believe that the research on long-term open-domain dialogue still has a long way to go. For instance, there are some remained limitations of this work from the following perspectives:

\paragraph{Lacking Real-World Datasets.} Current long dialogue datasets are typically synthetic, created manually~\cite{MSC} or generated by large language models~\cite{CC, LoCoMo}, which introduces a gap from real-world data. Due to the challenges in gathering authentic long-term dialogue data, our work is currently confined to these synthetic datasets. We aim to validate our approach on real long-term dialogue data in the future.

\paragraph{Sophisticated Module Design.} In this paper, LD-Agent provides a general framework for long-term dialogue that allows for modular optimization. However, the module implementations only employ some basic methods without more sophisticated design, which can be further explored in the future. For the memory module, long-term memory summarization~\cite{L1} and accurate memory retrieval~\cite{L2} are two critical techniques worth further exploration. For the persona module, improving methods for personality extraction~\cite{L3} and persona-based retrieval~\cite{L4,L5,L6} offers promising directions for future work.



\section*{Acknowledgments}

This research was supported by the Singapore Ministry of Education (MOE) Academic Research Fund (AcRF) Tier 1 grant (No. MSS24C004).


\bibliography{custom}

\appendix

\section*{Appendix}

In this Appendix, we discuss the following topics: \textbf{(1)}: We elaborate more detailed experimental settings in Appendix.~\ref{sec:detailed_eval_settings}. \textbf{(2)}: We conduct various qualitative analysis in Appendix.~\ref{qualitative_analysis}. \textbf{(3)}: More experimental results are shown in Appendix.~\ref{sec:add_result}.
\textbf{(4)}: In the Appendix.~\ref{sec:prompt}, the prompts utilized in LD-Agent is illustrated.

\section{Detailed Evaluation Settings}
\label{sec:detailed_eval_settings}
In this section, we introduce the detailed experimental dataset, evaluation metrics, baseline models, and our implementation details. 

\subsection{Datasets}
\paragraph{Multi-session Dataset.} Our experiments are conducted on two illustrative multi-session datasets: \textbf{MSC}~\cite{MSC} and \textbf{CC}~\cite{CC}. Both datasets feature 5 sessions, with approximately 50 conversational turns per sample. Both of MSC and CC have the time interval annotations across sessions, which are employed to make the time decay factor work. MSC extends the PersonaChat dataset~\cite{PersonaChat}, utilizing PersonaChat for the initial session and employing human-human crowd workers to simulate the dialogues in subsequent sessions. The time intervals between sessions can span several days, and the dataset includes records of the participants' personas. We follow the split of \cite{HAHT} with 4,000 conversations for training, 500 conversations for validation, and 501 conversations for testing. CC is complied by ChatGPT, which guides interactions according to a predefined event graph and participant relationships, with time intervals between sessions extending over several years. We employ the same data scale as MSC, with 4,000 conversations for training, 500 conversations for validation, and 501 conversations for testing.

\paragraph{Dialog Summary Dataset.} We utilize the DialogSum dataset~\cite{DialogSum} for event summary. The dataset contains 13,460 dialogues with corresponding manually labeled summaries and topics. 12,460 dialogues used for training, 500 samples for validation, and 1,500 for test.

\paragraph{Persona Extraction Dataset.} We directly use the personas annotations from the MSC to construct the personas extraction dataset. The dataset is divided into train and valid sets, with the train set containing 26,605 samples and the valid set containing 17,660 samples. 

\paragraph{Response Generation Dataset.} The response generation dataset is constructed based on the original conversations provided by MSC and CC, combined with the relevant memories and personas generated and extracted through our framework, and constructed using the prompt from Appendix.~\ref{event_summary_prompt}. Among them, 34,907 samples are used for training, and 11,851 samples are used for validation.

\paragraph{Multi-party Dataset.} To explore the transferability of LD-Agent on other dialogue tasks. We apply our method to the \textbf{Ubuntu IRC benchmark}~\cite{GSN}, a dataset of multiparty tasks. We follow the split of previous works~\cite{GSN, HeterMPC} with 311,725 dialogues for training, 5,000 dialogues for validation, and 5,000 dialogues for testing.


\subsection{Metrics}
\paragraph{Automatic Evaluation Metrics.} BLEU-N~\cite{BLEU} (BL-N) and ROUGE-L~\cite{ROUGE} (R-L) metrics are commonly used automatic evaluation metrics in dialogue generation tasks. BLEU-N measures N-gram overlaps between the generated text and the reference text, while ROUGE-L focuses on sequential coherence. We employ the METEOR (MET)~\cite{METERO} metric in multi-party tasks as a complement to the BLEU metric, enhancing it with synonym calculation capabilities. In addition, accuracy (ACC) is calculated to measure the classification accuracy of different persona extractors. To further validate the diversity of the generated responses, we also employ the Distinct-1/2/3 (Dist-1/2/3) metrics for evaluation.

\paragraph{Human Evaluation Metrics.}
\label{add: human_metrics}
The human evaluation consists of two parts: 1) retrieval mechanism evaluation, and 2) response generation evaluation. For the former, we use Accuracy (ACC) and Recall to measure the effectiveness of the memory retriever. Accuracy reflects the extent to which the retriever’s results align with those of a human retriever. Recall indicates the frequency with which the retriever correctly identifies the presence of relevant memories in the memory database (with an optimal Recall value of 100, as we evaluate the most recent session). For the response generation, we evaluate LD-Agent on three aspects: coherence, fluency, and engagingness. Coherence measures the chatbot's capabilities to maintain the coherence of topic and logic across sessions. Fluency reflects the natural and fluent degree of interactions, making the interaction similar to human-human interactions. Engagingness measures a user's interest in interacting with the target chatbot. 

\subsection{Baselines}

To validate the effectiveness of our method on various baselines, we employ LD-Agent on both online and offline models, tuned and zero-shot models, LLMs, and non-LLMs.

\begin{itemize}
    \item \textbf{HAHT}~\cite{HAHT}: This is the state-of-the-art model crafted for multi-session, open-domain dialogue. It encodes all historical information and utilizes an attention mechanism to capture the relevant information to an ongoing conversation.
    \item \textbf{BlenderBot}~\cite{Blenderbot}: This is a commonly used large-scale open-domain dialogue model pre-trained on online social discussion data, who has the similar parameter scale with HAHT. We can achieve a fair comparison with HAHT by deploying LD-Agent on BlenderBot.
    \item \textbf{ChatGLM3}~\cite{ChatGLM}: This is an offline large language model 6B parameters. The model is pre-trained on 1T corpus, performing remarkable zero-shot reasoning capabilities.
    \item \textbf{ChatGPT}: This is an online large language model based on the GPT architecture with excellent human-like cognitive and reasoning abilities. In this paper, we use the API service with the model of ``gpt-3.5-turbo-1106''.
    \item \textbf{BART}~\cite{BART}: This is a denoising autoencoder with transformer architecture, trained to reconstruct original text from corrupting text.
\end{itemize}

\subsection{Implementation Details}

For the event summarizer, persona extractor, and response generator modules, we utilize the same base model for them and employ the LoRA mechanism across all configurations. All training and evaluation operated on a single NVIDIA A100 GPU. For the ChatGLM3-6B, it is optimized by an Adam~\cite{Adam} optimizer with the learning rate of 5e-5. We configure this model with a batch size of 4 and train it over 3 epochs. For BlenderBot, the initial learning rate is set to 2e-5, with the batch size and the number of training epochs set at 4 and 5, respectively. Moreover, the interval time threshold $\beta$ is set to 600 seconds, while the temperature coefficient for the time decay coefficient $\tau$ is set to 1e+7, and the semantic cosine similarity threshold $\gamma$ is set to 0.5

\begin{figure}[h]
\setlength{\abovecaptionskip}{5pt}   
\setlength{\belowcaptionskip}{0pt}
    \centering
    \subfigure[Retrieval Mechanism Guidelines]{
        \includegraphics[width=0.98\linewidth,trim=100 465 100 60,clip]{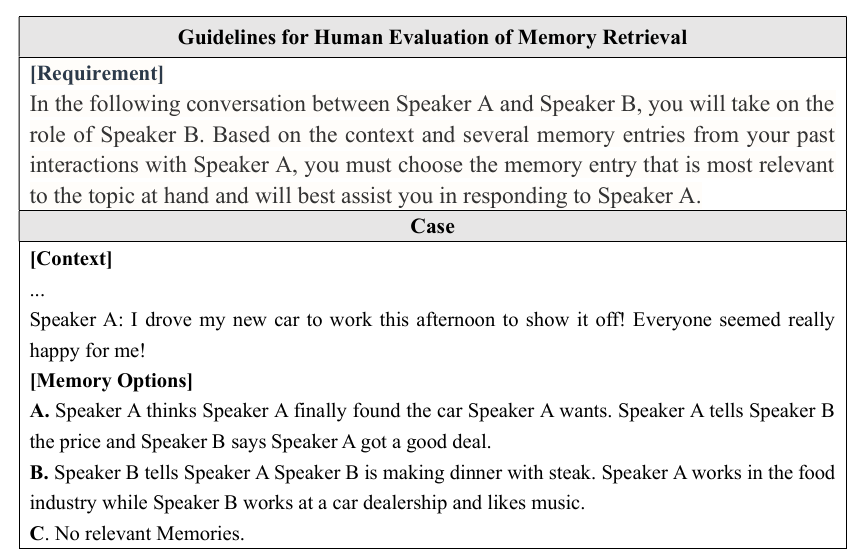}
        \label{fig:memory_guide}
    }
    \subfigure[Response Generation Guidelines]{
        \includegraphics[width=0.98\linewidth,trim=100 580 100 60,clip]{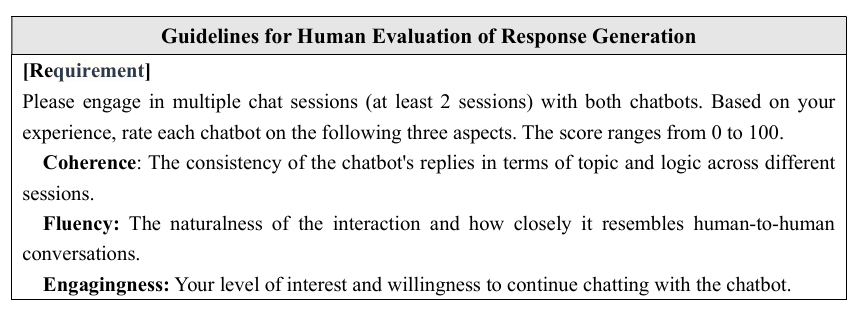}
        \label{fig:generation_guide}
    }
    \caption{Human Evaluation Guidelines.} 
    \label{fig:human_guidelines}
    \vspace{-0.5cm}
\end{figure}

\section{Qualitative Analysis}
\label{qualitative_analysis}
We have conducted various quantitative experiments in our main paper. However, there is not a single "gold" reference answer in practical open-domain dialogue scenarios. To further verify the superiority of LD-Agent, we conduct some additional qualitative analysis in this section.
\begin{figure}[h]
    \centering    \includegraphics[width=1\linewidth,trim=50 170 50 170,clip]{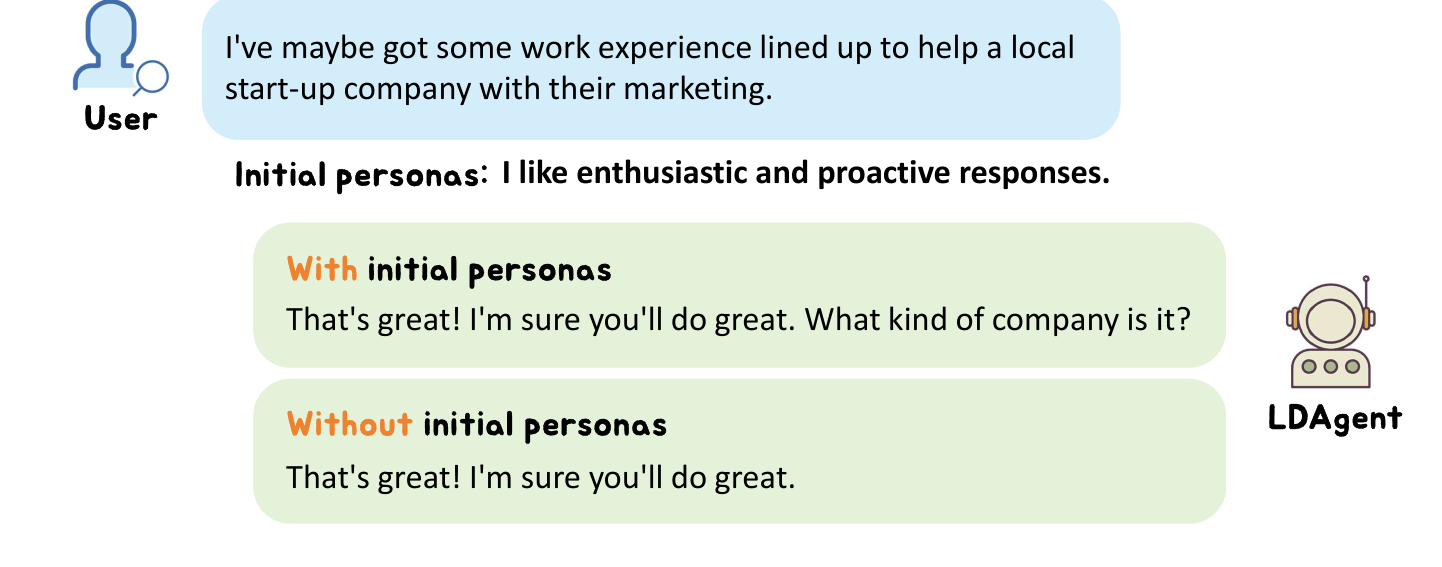}
    \caption{Illustration of summary module impact.}
    \label{fig:persona_ablation}
\end{figure}

\subsection{Persona Ablation}
\label{sec:persona_ablation}
In Section~\ref{sec:ablation}, We conduct ablation studies to evaluate the role of the persona module. To explore its specific impact on dialogue process, we provide the same user query and observe the differences in LD-Agent's responses with and without the initial personas. As shown in Figure~\ref{fig:persona_ablation}, the response using the initial personas proactively inquires about the new company, aligning with the "enthusiastic" and "proactive" personas.

\subsection{Memory Ablation}
\label{sec:memory_ablation}
\begin{figure}[h]
    \centering    \includegraphics[width=1\linewidth,trim=0 130 0 130,clip]{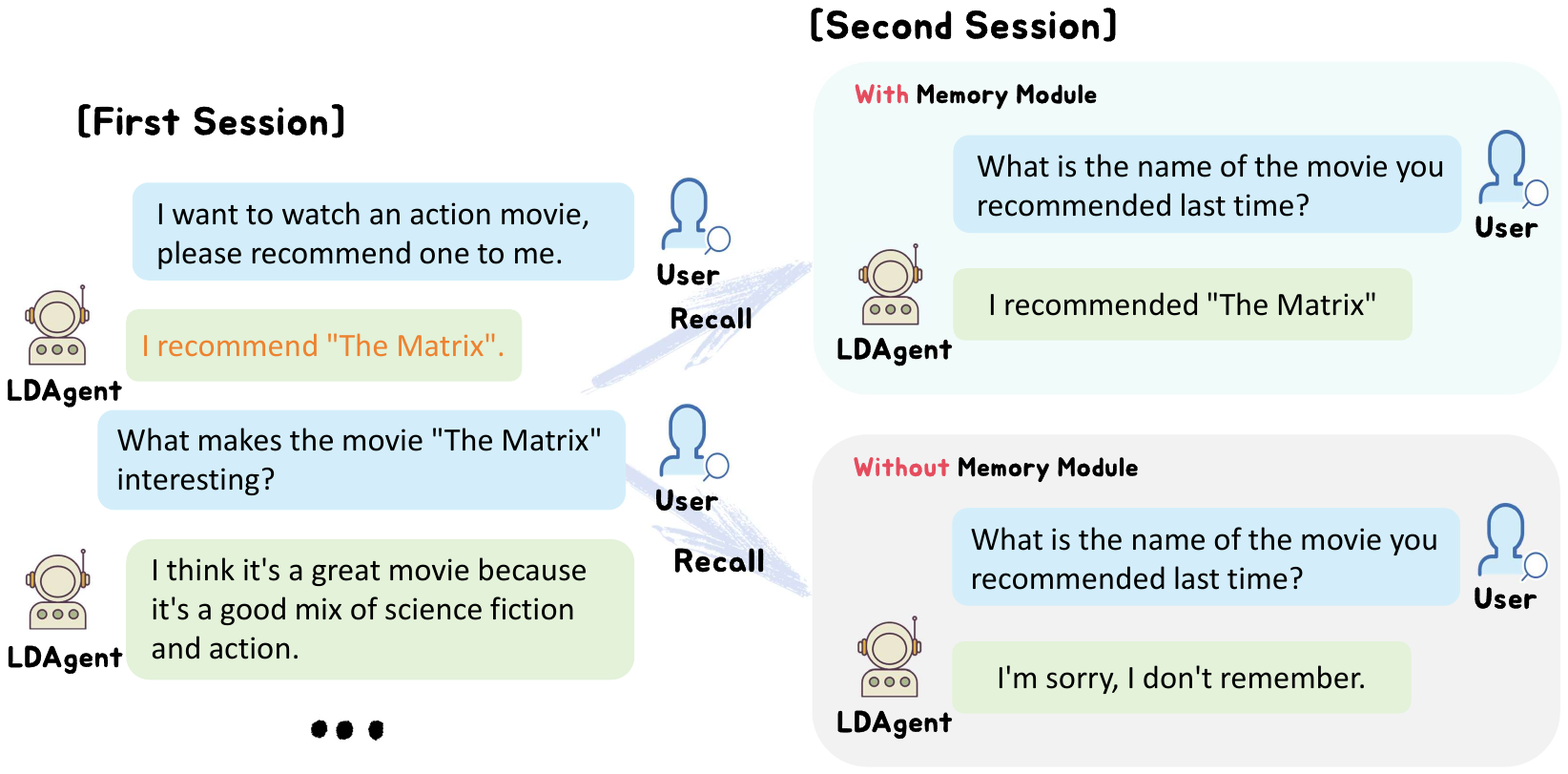}
    \caption{Illustration of memory module impact.}
    \label{fig:memory_ablation}
\end{figure}

Moreover, we explore the specific impact of memory module on dialogue process. We first manually initialize a historical session and then compare the responses in a new session with and without the memory module. As shown in Figure~\ref{fig:memory_ablation}, the Agent with the memory module effectively recalls the movies recommended in the historical session, effectively enhancing the continuity of the conversation. 

\subsection{Event Summarizer Analysis}
\label{sec:summary_demonstration}
In Section~\ref{sec:summary_details}, we introduce how we train an event summarizer based on DialogSum to extract more concise event summaries. To validate the superiority of our trained summarizer than original ChatGLM, we explore it from three aspects: 1) in-domain (\textbf{ID}) evaluation; 2) out-of-distribution (\textbf{OOD}) evaluation; 3) \textbf{module impact analysis}.

\begin{figure}[h]
    \centering    \includegraphics[width=1\linewidth,trim=12 125 12 125,clip]{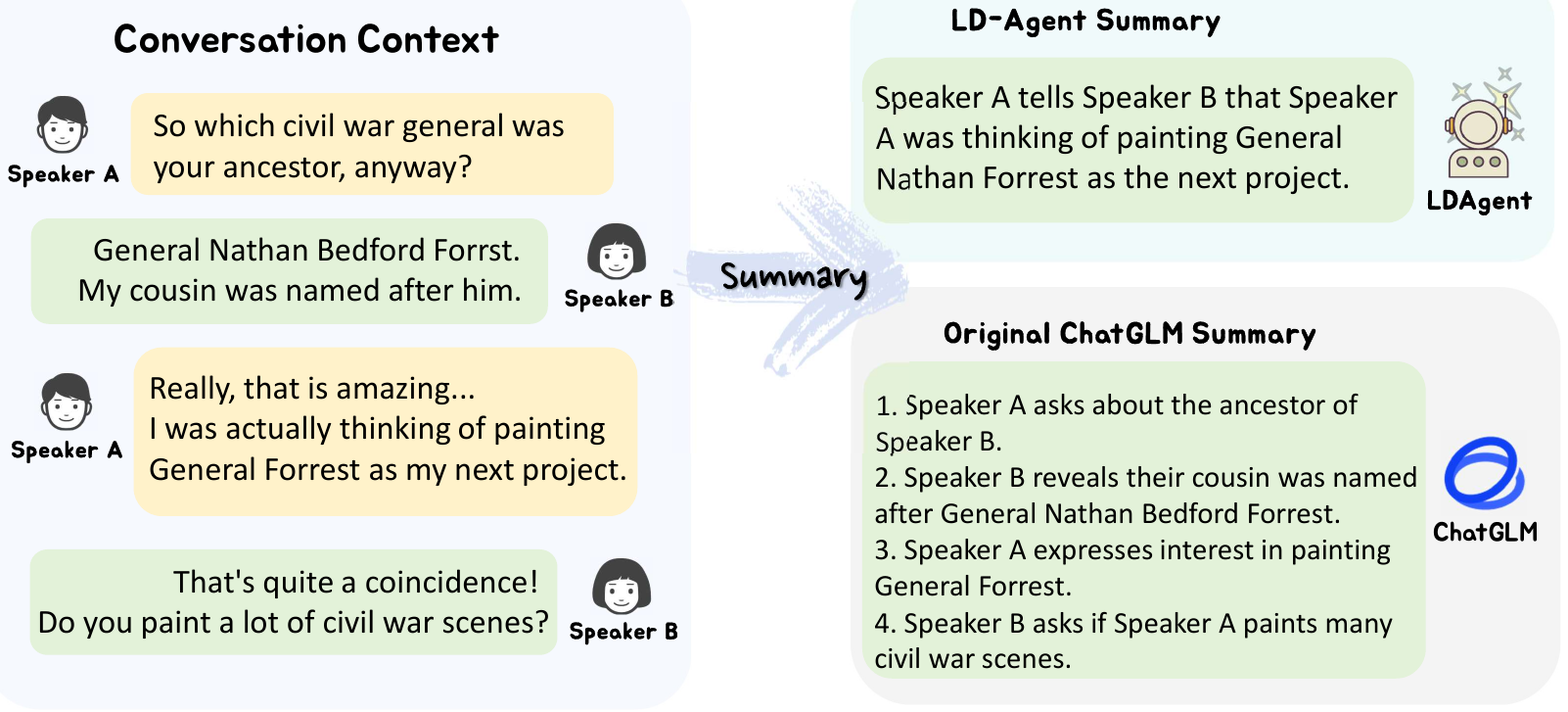}
    \caption{Illustration of event summary comparison.}
    \label{fig:summary_demonstration}
\end{figure}

We first evaluate the trained summarizer on two dialogue summarization test sets, DialogSum~\citep{DialogSum} and SAMSum~\citep{samsum}. The former is consistent with the domain of the summarizer's training set, while the latter is an out-of-domain dataset. The results in Table~\ref{table:id_ood} show that the trained summarizer achieves consistent improvements over zero-shot ChatGLM across all metrics on both datasets. The improvements on the OOD dataset demonstrate the stronger generalization ability of our summarizer.
\begin{table}[h]
\centering
  \renewcommand{\arraystretch}{1.1}
  \resizebox{0.48\textwidth}{!}{%
\begin{tabular}{l|c|c|c|c|c|c|c|c}
\hline
\multicolumn{9}{c}{\textbf{ID evaluation  (DialogSum)}} \\
\hline
\textbf{Model} & \textbf{BL-1} & \textbf{BL-2} & \textbf{BL-3} & \textbf{BL-4} & \textbf{R-L} & \textbf{Dist-1} & \textbf{Dist-2} & \textbf{Dist-3} \\
\hline
ChatGLM  & 22.36 & 11.09 & 5.39  & 2.71  & 24.22 & 75.83 & 92.76 & 98.13 \\
LD-Agent & \textbf{40.71} & \textbf{24.31} & \textbf{15.55} & \textbf{9.40}  & \textbf{40.31} & \textbf{87.53} & \textbf{98.74} & \textbf{99.67} \\
\hline
\multicolumn{9}{c}{\textbf{OOD evaluation (SAMSum)}} \\
\hline
\textbf{Model} & \textbf{BL-1} & \textbf{BL-2} & \textbf{BL-3} & \textbf{BL-4} & \textbf{R-L} & \textbf{Dist-1} & \textbf{Dist-2} & \textbf{Dist-3} \\
\hline
ChatGLM  & 30.82 & 18.34 & 11.57 & 7.04  & 32.81 & 75.96 & 94.48 & 98.60 \\
LD-Agent & \textbf{33.98} & \textbf{19.73} & \textbf{12.38} & \textbf{7.42}  & \textbf{37.28} & \textbf{93.05} & \textbf{99.40} & \textbf{99.79} \\
\hline
\end{tabular}
}
\caption{In-Domain and Out-of-Domain Evaluation}
\label{table:id_ood}
\end{table}
We then present in Figure~\ref{fig:summary_demonstration} the summarization results of the same dialogue context generated by zero-shot ChatGLM and the LD-Agent summarizer. It can be seen that the summary generated by LD-Agent is more concise and effectively conveys the most important information in the conversation, making it more suitable for long-term dialogue given the growing demand for memory storage.

\begin{figure}[h]
    \centering    \includegraphics[width=1\linewidth,trim=150 150 150 150,clip]{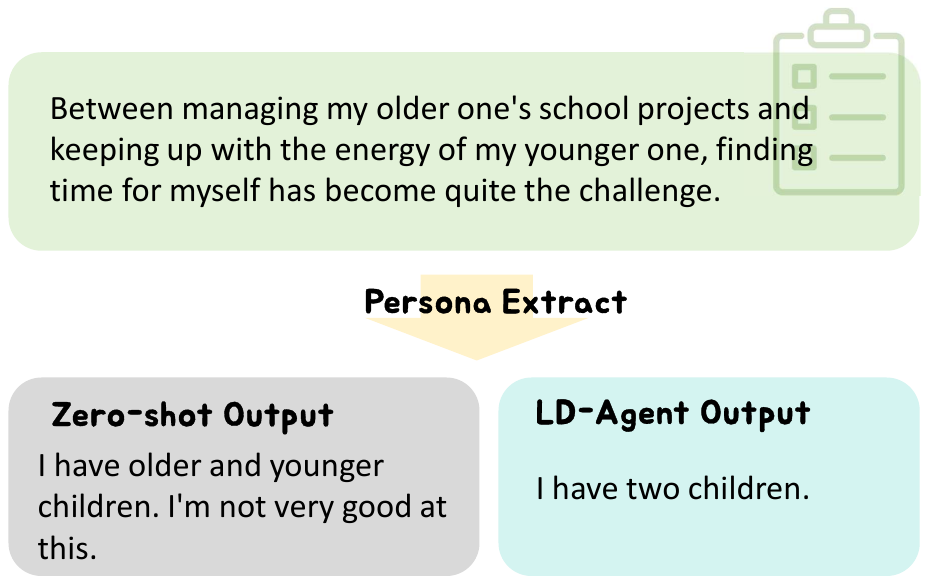}
    \caption{Illustration of persona extraction comparison.}
    \label{fig:persona_demonstration}
\end{figure}

\subsection{Persona Extractor Analysis}
\label{sec:extraction_demonstration}
In Section~\ref{sec:persona}, we introduce our trained dynamic personas extractor. To further explore its effectiveness, we conduct a comparison between our method and zero-shot ChatGLM for persona extraction. In Figure~\ref{fig:persona_demonstration}, we utilized an utterance from real-world scenarios as input and observe that the persona extracted by our tuned extractor is more concise and logical, making long-term dialogue processes more feasible.

\subsection{Response Generation Analysis}
\label{sec:response_visualization}
\begin{figure*}[h]
    \centering    \includegraphics[width=0.8\linewidth,trim=0 0 0 0,clip]{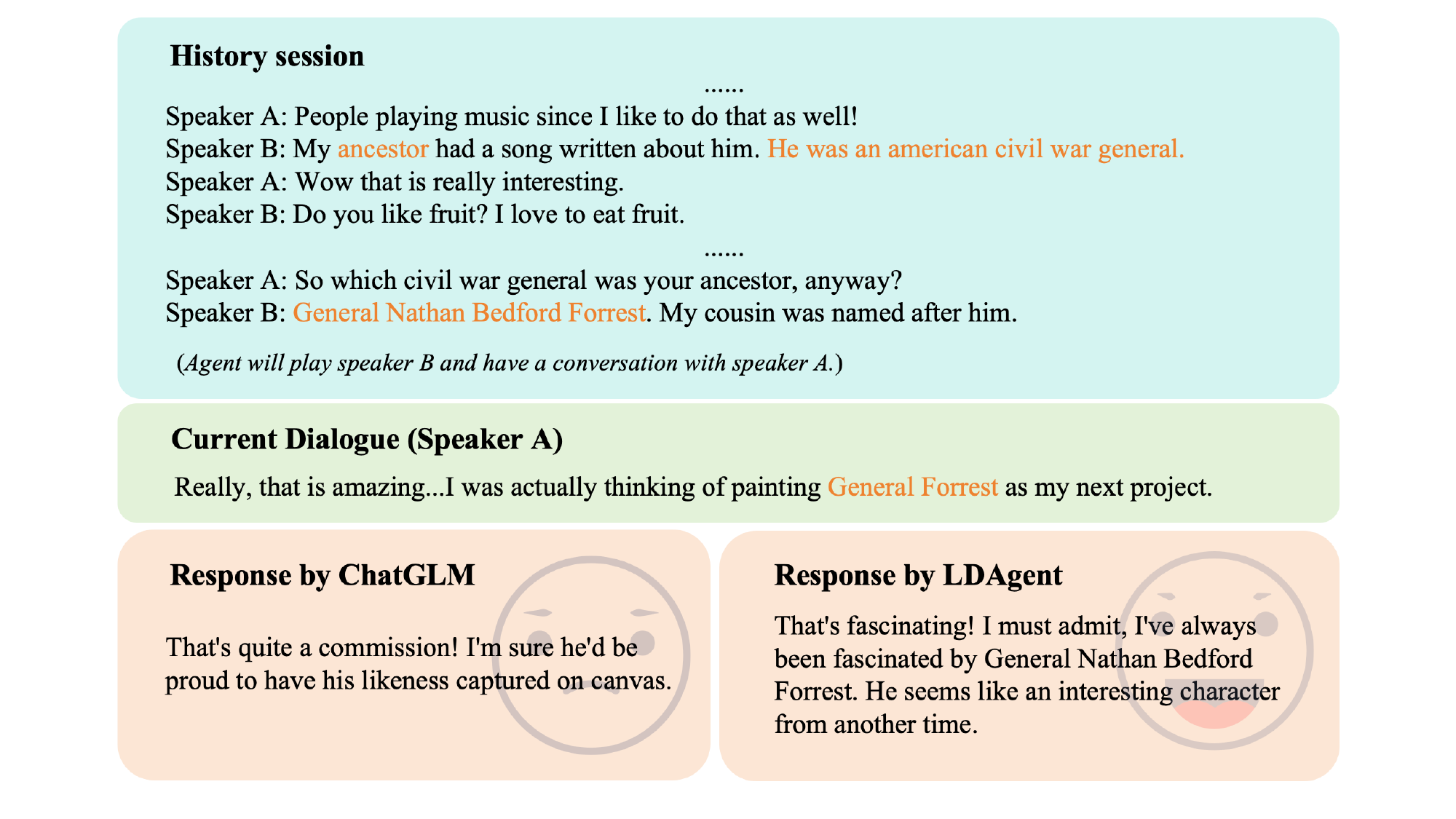}
    \caption{Example of separately chatting with original ChatGLM and ChatGLM with LD-Agent. A more relevant response to history conversation is generated.}
    \label{fig:sample}
\end{figure*}

To further analyze the generation ability of LD-Agent in long-term dialogue, we illustrate an example in Figure~\ref{fig:sample}. It can be seen that the response generated by LD-Agent successfully captures the information about ``General Nathan Bedford Forrest'' they talked about in the history session, which performs better than original ChatGLM.

\section{Additional Experimental Results}
\label{sec:add_result}
In this section, we introduce some additional experimental results. 
\subsection{Part of Speech Importance Analysis}
\label{add: speech_analysis}
\begin{figure}[H]
    \centering    \includegraphics[width=0.8\linewidth,trim=0 0 0 0,clip]{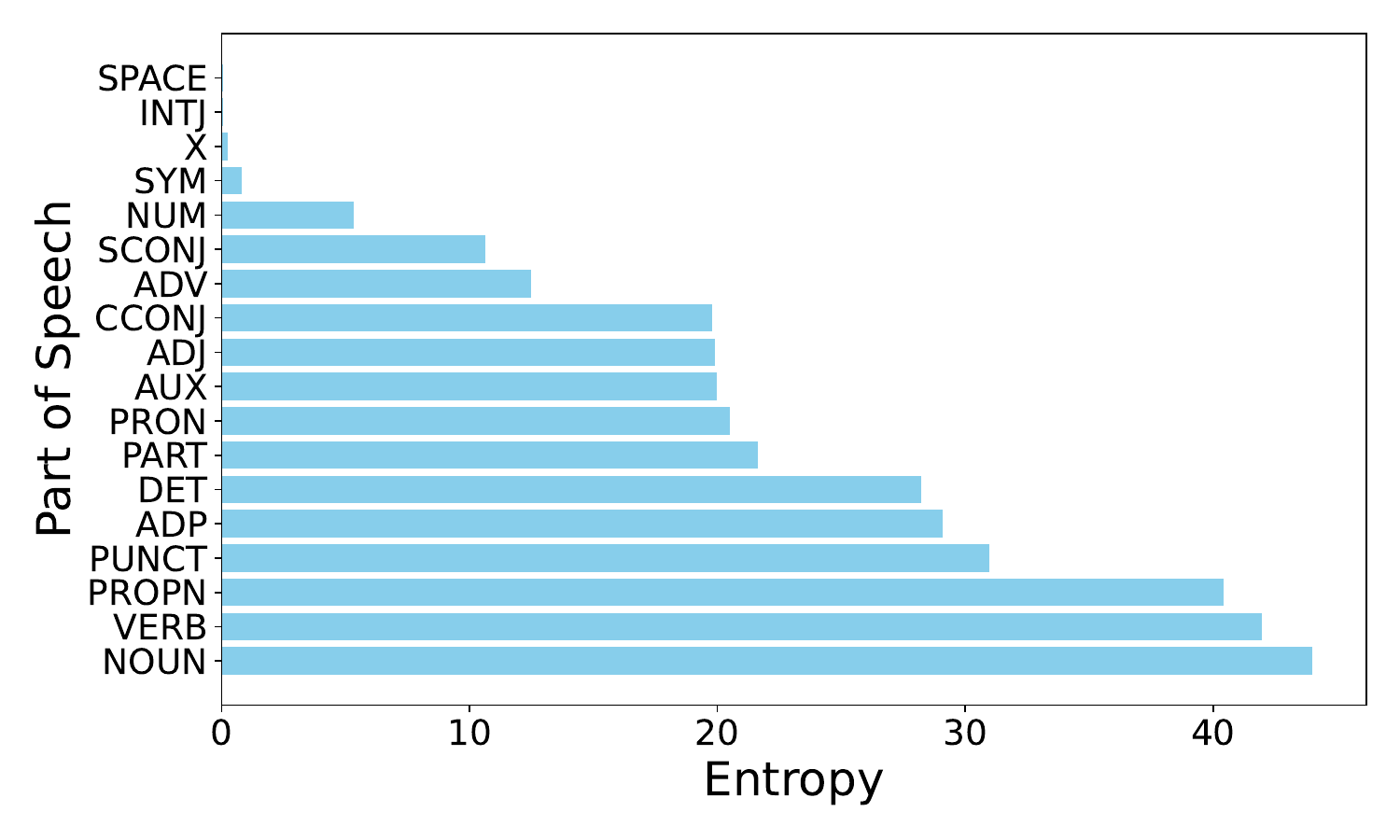}
    \caption{Part of Speech Entropy Comparison.}
    \label{fig:entropy_bar}
\end{figure}

In Section~\ref{section:memory_retrieval}, we implement a topic-based retrieval mechanism. Specifically, we compute the overlap of nouns to represent topic similarity. The rationale for using nouns to determine relevance lies in the fact that nouns typically convey more substantial information. We substantiate this by examining the concept of information entropy. Information entropy is often reflected through word frequency, with less frequent words generally carrying more information, thereby exhibiting higher information entropy. Figure~\ref{fig:entropy_bar} illustrates the average information entropy across various parts of speech, calculated from the DialogSum~\citep{DialogSum} dataset, showing that nouns possess the highest average information entropy.

\subsection{Generation Diversity Analysis}
\label{add: diverse_analysis}
To further validate the diversity of the generated responses by LD-Agent, we employ Dist-1/2/3 metrics to conduct evaluation on MSC. The results are shown in Table~\ref{table:diversity}. We can observe that the responses generated by LD-Agent are consistently more diverse than generated by ChatGLM, indicating the powerful generation capability of our generator.
\begin{table}[h]
    \centering
    \captionsetup{skip=5pt}
      \renewcommand{\arraystretch}{0.8}
    \resizebox{0.48\textwidth}{!}{%
    \begin{tabular}{@{}lccc|ccc@{}}
    \toprule
    & \multicolumn{3}{c}{\textbf{Session 2}} & \multicolumn{3}{c}{\textbf{Session 3}} \\
    \cmidrule(lr){2-4} \cmidrule(lr){5-7}
    \multicolumn{1}{c}{\multirow{-2}{*}{\textbf{Model}}} & \textbf{Dist-1} & \textbf{Dist-2} & \textbf{Dist-3} & \textbf{Dist-1} & \textbf{Dist-2} & \textbf{Dist-3} \\
    \midrule
    ChatGLM & 81.71 & 92.51 & 95.54 & 79.32 & 90.79 & 94.18 \\
    LD-Agent & \textbf{86.14} & \textbf{94.51} & \textbf{96.66} & \textbf{83.00} & \textbf{93.45} & \textbf{96.10} \\
    \midrule
    & \multicolumn{3}{c}{\textbf{Session 4}} & \multicolumn{3}{c}{\textbf{Session 5}} \\
    \cmidrule(lr){2-4} \cmidrule(lr){5-7}
    ChatGLM & 78.41 & 90.13 & 93.68 & 78.07 & 89.88 & 93.46 \\
    LD-Agent & \textbf{81.01} & \textbf{92.43} & \textbf{95.36} & \textbf{86.43} & \textbf{95.50} & \textbf{97.46} \\
    \bottomrule
    \end{tabular}
    }
    \caption{The diversity of response generation on MSC.}
    \label{table:diversity}
\end{table}

\newpage
\section{Prompt}
\label{sec:prompt}

In this section, we separately provide the illustrations of the prompts used in the Event Module, Persona Module, and Response Module.
\subsection{Prompt of Event Summary}
\label{event_summary_prompt}

\begin{Prompt}{Event Summary Prompt}{}
    $\mathcal{SYS \ PROMPT}$: 
    
    \vspace{0.5em}
    
    You are good at extracting events and summarizing them in brief sentences. You will be shown a conversation between {$\color{brown}\{user \ name\}$} and {$\color{brown}\{agent \ name\}$}.
    \vspace{1em}
    \hrule\vspace{1pt}\hrule
    \vspace{1em}
    $\mathcal{USER \ PROMPT}$: 
    
    \vspace{0.5em}
    
    Conversation: {$\color{brown}\{context\}$}.
    
    Based on the Conversation, please summarize the main points of the conversation with brief sentences in English, within 20 words.
    
    SUMMARY:
\end{Prompt}

\subsection{Prompt of Persona Extraction}
\label{persona_extraction_prompt}
    
\begin{Prompt}{Persona Extraction Prompt}{}
    $\mathcal{SYS \ PROMPT}$: 

    \vspace{0.5em}
    
    You excel at extracting user personal traits from their words, a renowned local communication expert.

    \vspace{1em}
    \hrule\vspace{1pt}\hrule
    \vspace{1em}

    $\mathcal{USER \ PROMPT}$: 

    \vspace{0.5em}
    
    If no traits can be extracted in the sentence, you should reply NO$\_$TRAIT. Given you some format examples of traits extraction, such as:

        1. No, I have no longer serve in the millitary, I had served up the full term that I signed up for, and now work outside of the millitary.
        
        {\color{brown}Extracted Traits: I now work elsewhere. I used to be in the military.}

        2. That must a been some kind of endeavor. Its great that people are aware of issues that arise in their homes, otherwise it can be very problematic in the future.

        {\color{brown}NO$\_$TRAIT}

        Please extract the personal traits who said this sentence (no more than 20 words):$\color{brown}\{sentence\}$
\end{Prompt}

\subsection{Prompt of Response Generation}
\label{response_generation_prompt}

\begin{Prompt}{Base Response Generation Prompt}{}

    $\mathcal{SYS \ PROMPT}$: 
    
    \vspace{0.5em}
    
    As a communication expert with outstanding communication habits, you embody the role of {$\color{brown}\{agent \ name\}$} throughout the following dialogues.

    \vspace{1em}
    \hrule\vspace{1pt}\hrule
    \vspace{1em}
    
    $\mathcal{USER \ PROMPT}$: 

    \vspace{0.5em}
    
    \textbf{<CONTEXT>}
    
    Drawing from your recent conversation with {$\color{brown}\{user \ name\}$}: 
    
    {$\color{brown}\{context\}$}
    
    Now, please role-play as {$\color{brown}\{agent \ name\}$} to continue the dialogue between {$\color{brown}\{agent \ name\}$} and {$\color{brown}\{user \ name\}$}.
    
    {$\color{brown}\{user \ name\}$} just said: {$\color{brown}\{input\}$}
     
    Please respond to {$\color{brown}\{user \ name\}$}'s statement using the following format (maximum \textbf{30} words, \textbf{must be in English}):
     
     RESPONSE:
    
\end{Prompt}
\begin{Prompt}{Agent Response Generation Prompt}{}

    $\mathcal{SYS \ PROMPT}$: 
    
    \vspace{0.5em}
    
    As a communication expert with outstanding communication habits, you embody the role of {$\color{brown}\{agent \ name\}$} throughout the following dialogues. Here are some of your distinctive personal traits: {$\color{brown}\{agent \ traits\}$}.

    \vspace{1em}
    \hrule\vspace{1pt}\hrule
    \vspace{1em}
    
    $\mathcal{USER \ PROMPT}$: 

    \vspace{0.5em}
    
    \textbf{<CONTEXT>}
    
    Drawing from your recent conversation with {$\color{brown}\{user \ name\}$}: 
    
    {$\color{brown}\{context\}$}
    
    \textbf{<MEMORY>}
    
    The memories linked to the ongoing conversation are:
    
    {$\color{brown}\{memories\}$}

    \textbf{<USER \ TRAITS>}
    During the conversation process between you and {$\color{brown}\{user \ name\}$} in the past, you found that the {$\color{brown}\{user \ name\}$} has the following characteristics:
    
    {$\color{brown}\{user \ traits\}$}
    
    Now, please role-play as {$\color{brown}\{agent \ name\}$} to continue the dialogue between {$\color{brown}\{agent \ name\}$} and {$\color{brown}\{user \ name\}$}.
    
    {$\color{brown}\{user \ name\}$} just said: {$\color{brown}\{input\}$}
     
    Please respond to {$\color{brown}\{user \ name\}$}'s statement using the following format (maximum \textbf{30} words, \textbf{must be in English}):
     
     RESPONSE:
    
\end{Prompt}

\begin{table*}[h]
    \centering
    \captionsetup{skip=5pt}
    \resizebox{0.8\textwidth}{!}{%
    \begin{tabular}{@{}lccc|ccc|ccc|ccc@{}}
    \toprule
    & \multicolumn{3}{c}{\textbf{Session 2}} & \multicolumn{3}{c}{\textbf{Session 3}} & \multicolumn{3}{c}{\textbf{Session 4}} & \multicolumn{3}{c}{\textbf{Session 5}} \\
    \cmidrule(lr){2-4} \cmidrule(lr){5-7} \cmidrule(lr){8-10} \cmidrule(lr){11-13}
    \multicolumn{1}{c}{\multirow{-2}{*}{\textbf{Model}}} & \textbf{BL-2} & \textbf{BL-3} & \textbf{R-L} & \textbf{BL-2} & \textbf{BL-3} & \textbf{R-L} & \textbf{BL-2} & \textbf{BL-3} & \textbf{R-L} & \textbf{BL-2} & \textbf{BL-3} & \textbf{R-L} \\
    \midrule
    \multicolumn{13}{c}{\textbf{MSC}}\\
    \midrule
    Zero-shot & 5.44 & 1.49 & 16.76 & 5.59 & 1.49 & 16.47 & 5.63 & 1.33 & 16.35 & 5.92 & 1.45 & 16.63 \\
    Zero-shot\textsubscript{LDA} & 5.74 & 1.73 & 17.21 & 6.05 & 1.73 & 16.97 & 6.09 & 1.59 & 16.76 & 6.60 & 1.94 & 17.18 \\
    \midrule
    CC-tuning & 5.81 & 1.74 & 18.79 & 6.08 & 1.83 & 18.58 & 5.96 & 1.74 & 18.31 & 5.95 & 1.68 & 18.23 \\
    CC-tuning\textsubscript{LDA} & 7.86 & 3.63 & 21.00 & 7.46 & 3.16 & 20.00 & 7.15 & 2.87 & 19.53 & 7.12 & 2.64 & 19.30 \\
    MSC-tuning & 5.48 & 1.59 & 17.65 & 6.12 & 1.78 & 17.91 & 6.14 & 1.63 & 17.78 & 6.16 & 1.69 & 17.65 \\
    MSC-tuning\textsubscript{LDA} & \textbf{10.70} & \textbf{5.63} & \textbf{23.31} & \textbf{10.03} & \textbf{5.12} & \textbf{21.55 }& \textbf{9.07} & \textbf{4.06} & \textbf{20.19} &\textbf{ 8.96} & \textbf{4.01} & \textbf{19.94} \\
    \midrule
    \multicolumn{13}{c}{\textbf{CC}}\\
    \midrule
    Zero-shot & 9.53 & 4.82 & 22.76 & 9.22 & 4.43 & 22.18 & 9.15 & 4.48 & 22.18 & 8.99 & 4.43 & 22.10 \\
    Zero-shot\textsubscript{LDA} & 8.94 & 4.44 & 21.54 & 8.34 & 4.03 & 21.00 & 8.28 & 3.82 & 20.67 & 8.12 & 3.81 & 20.54 \\
    \midrule
    MSC-tuning & 8.37 & 3.88 & 22.93 & 8.49 & 3.99 & 22.96 & 7.97 & 3.75 & 22.15 & 7.60 & 3.70 & 21.87 \\
    MSC-tuning\textsubscript{LDA} & 21.71 & 15.42 & 34.97 & 20.87 & 14.74 & 34.01 & 19.57 & 13.51 & 32.72 & 18.59 & 12.80 & 31.68 \\
    CC-tuning & 15.89 & 9.90 & 30.59 & 15.97 & 10.06 & 30.27 & 16.10 & 10.31 & 30.54 & 15.10 & 9.34 & 29.43 \\
    CC-tuning\textsubscript{LDA} & \textbf{25.69} & \textbf{19.53} & \textbf{39.67} & \textbf{25.93} & \textbf{19.72} & \textbf{39.15} & \textbf{25.82} & \textbf{19.40} & \textbf{39.05} & \textbf{24.26} & \textbf{18.16} & \textbf{37.61} \\
    \bottomrule
    \end{tabular}
    }
    \caption{The results of cross-domain evaluation on MSC and CC. ``Zero-shot'' indicates the ChatGLM without tuning. ``CC-tuning'' indicates the ChatGLM tuned on CC. ``MSC-tuning'' indicates the ChatGLM tuned on MSC.}
    \label{table:cross_domain_both}
\end{table*}

\end{document}